\documentclass{article}
\usepackage{spconf,amsmath,amssymb,amsthm,graphicx,booktabs,multirow}
\usepackage{tikz}
\usetikzlibrary{arrows.meta,positioning,calc,fit,backgrounds}
\usepackage{pgfplots}
\pgfplotsset{compat=1.18}
\usepackage{microtype,balance}
\usepackage{hyperref}
\usepackage{graphicx}
\usepackage{xcolor}
\title{PHYSICS-INFORMED BUT NOT PHYSICS-CONSISTENT:\\
ERROR GEOMETRY AND SUBSPACE PROJECTION FOR NEURAL AC POWER FLOW}

\name{\begin{tabular}{c}
Changhun Kim$^{1}$, Timon Conrad$^{2}$, Redwanul Karim$^{1}$, Karan Pahlajani$^{1}$, Julian Oelhaf$^{1}$, David Riebesel$^{2}$\\
Tom\'as Arias-Vergara$^{1}$, Andreas Maier$^{1}$, Johann J\"ager$^{2}$, Siming Bayer$^{1}$
\end{tabular}%
\thanks{Corresponding author: changhun.kim@fau.de. This work was supported by the GridAssist project under the ``OptiNetD'' initiative of the German Federal Ministry for Economic Affairs and Energy (BMWE) and by HPC resources from NHR@FAU, partially funded by the German Research Foundation (DFG).}
}
\address{$^{1}$Pattern Recognition Lab, Friedrich-Alexander-Universit\"at Erlangen-N\"urnberg, Germany\\
$^{2}$Institute of Electrical Energy Systems, Friedrich-Alexander-Universit\"at Erlangen-N\"urnberg, Germany}

\begin{document}
\ninept
\maketitle

\begin{abstract}
Recent neural power-flow solvers, including emerging foundation models,
achieve accurate voltage predictions, yet such accuracy does not
necessarily imply physically consistent solutions.
Even small complex voltage errors can yield large AC power-balance residuals.
We study this accuracy--consistency gap across PIGNN-GC, GridSFM,
gridfm-graphkit, and LUMINA on realistic 2224-bus Great Britain network (GBnetwork) scenarios,
with cross-grid evaluation of GridSFM over 31 systems.
Using a singular value decomposition (SVD) basis fitted to training AC power-flow solutions, we find
that neural prediction errors contain substantial components outside the
dominant solution subspace. To address this mismatch, calibrated
solution-subspace projection (CSP) suppresses off-subspace prediction
components after train-only bias calibration, reducing Mean PB by $67.0\%$, $37.8\%$, $40.5\%$, and $68.9\%$
for PIGNN-GC, GridSFM, gridfm-graphkit, and LUMINA, respectively,
relative to calibrated predictions, while improving voltage-magnitude
accuracy in all four models.
These results identify output-error geometry as an important factor in
physics-consistent neural AC power flow.
Code: \href{http://github.com/Kimchangheon/neural-acpf-error-geometry}{github.com/Kimchangheon/neural-acpf-error-geometry}
\end{abstract}


\section{Introduction}
\label{sec:introduction}

AC power flow (AC-PF) is a fundamental tool for steady-state power-system
analysis \cite{zimmerman2011matpower}. It is repeatedly solved in congestion
management \cite{conrad2026congestion} and contingency studies
\cite{bolz2023contingency}, and underlies optimal power flow (OPF)
\cite{frank2012opf,owerko2020opf} and state-estimation workflows
\cite{cheng2024stateestimation}.
The need for both computational efficiency and physical consistency has
motivated neural surrogates including direct data-driven approximation~\cite{donon2020graphsolver,lin2024powerflownet}, physics-informed training \cite{owerko2024unsupervised,jiang2025upinn,wen2026pigat}, feasibility-seeking mechanisms \cite{nguyen2025fsnet,donti2021dc3}, and learned projection-based correction \cite{chzhen2026fab}.

More recent cross-grid and
foundation-model efforts--including GridSFM \cite{yang2026gridsfm},
gridfm-graphkit and GENCO \cite{gridfmgraphkit2026,puech2026genco},
and LUMINA \cite{li2026lumina}--extend this direction with global context
and topology-aware representations of heterogeneous grid components \cite{dejongh2022physics} designed to operate
across diverse grids and operating conditions \cite{karim2026domainadaptation}.

However, accurate voltage prediction does not necessarily imply physical
consistency \cite{ma2026safepowergraph,shamseldein2026hybrid}.
Physics-informed objectives can reduce power-flow residuals
\cite{nellikkath2022pinn,hu2021physicsguided}, but remain soft training
objectives and do not by themselves enforce exact satisfaction of the nonlinear
AC equations \cite{kim2026pignn}. Consequently, neural surrogates with small
voltage errors can still exhibit non-negligible power-balance violations.
We refer to this discrepancy as the \emph{accuracy--consistency gap} and ask
which properties of the remaining prediction error govern its physical impact.

We hypothesize that error \emph{direction}, rather than magnitude alone, is
an important factor: similarly sized voltage errors can induce substantially
different AC residuals depending on their orientation across buses. Recent work has examined directional effects of warm-start errors in the distinct
context of Newton--Raphson (NR) convergence~\cite{bose2026newtons}. This
motivates studying the \emph{geometry} of neural-surrogate errors in the
voltage-state space.

\begin{figure}[t]
    \centering
    \includegraphics[width=\columnwidth]
    {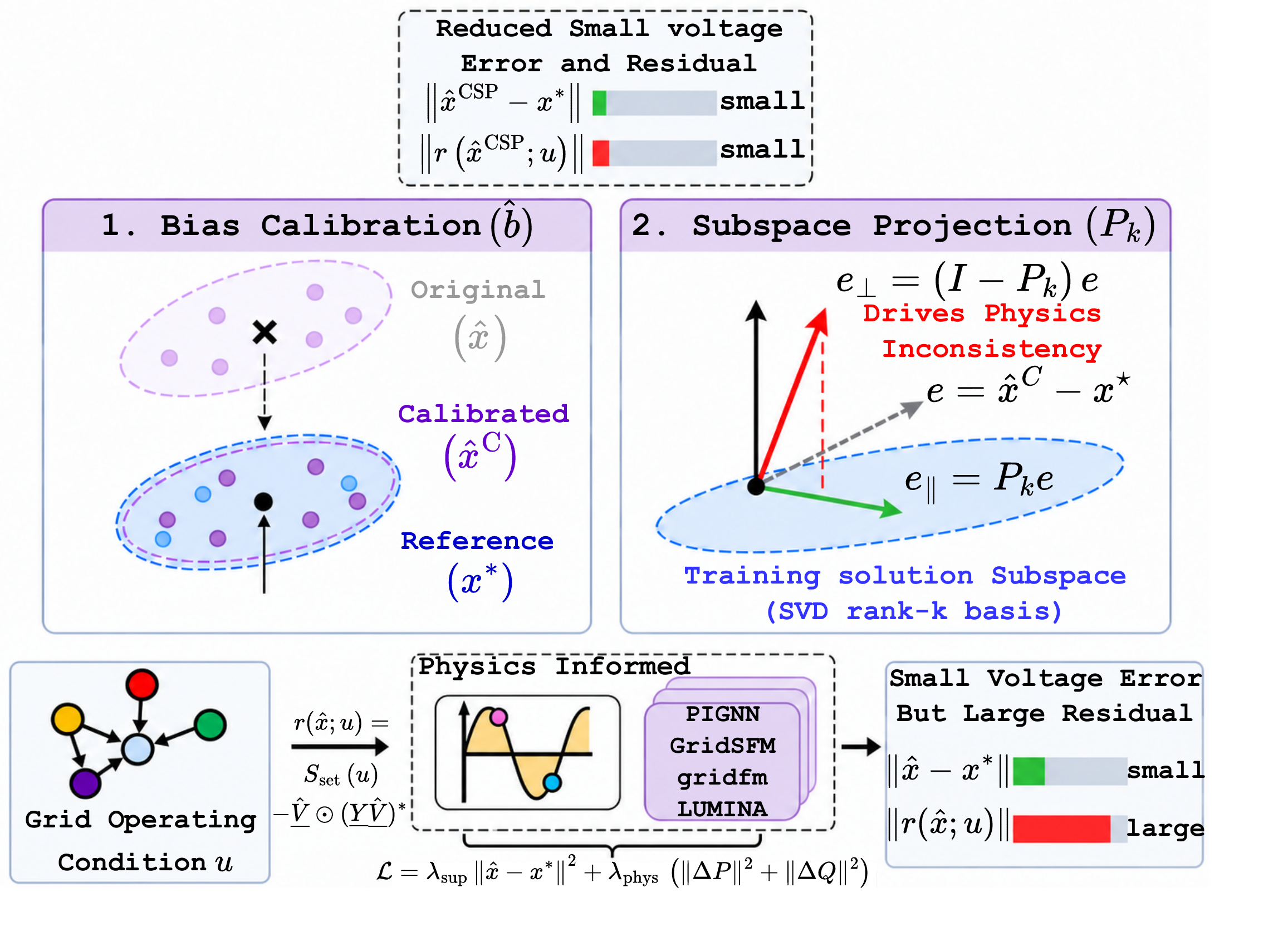}
    \caption{A low-dimensional subspace is derived from the training solutions. CSP first removes train-estimated bias and projects the corrected prediction onto this subspace, suppressing off-subspace errors and improving voltage accuracy and AC power-balance consistency.}
    \label{fig:graphical_abstract}
\end{figure}

Goodwin et al.~\cite{goodwin2026geometry} characterized the differential
geometry of the AC-PF solution manifold, while
Park et al.~\cite{park2024compact} and Chevalier et al.~\cite{chevalier2022accelerated}
demonstrated low-dimensional structure in AC-OPF and PF solutions,
respectively. Motivated by this evidence of low-dimensional solution
structure, we ask whether the relationship between neural prediction
errors and a \emph{training solution subspace} can be quantified to explain
physical inconsistency, and whether this structure can be exploited to
reduce power-balance violations.

To address these questions, we make three contributions.
(i) We decompose calibrated prediction errors of four neural surrogates
into in- and off-subspace components using a low-rank singular value
decomposition (SVD) basis fitted to training AC-PF solutions, and show
through fixed-norm interventions and local Jacobian analysis that
off-subspace errors are substantially more residual-sensitive.
(ii) Building on this finding, we propose \emph{calibrated
solution-subspace projection} (CSP), a lightweight post-hoc output correction that
uses only training data, removes systematic bias, and suppresses
off-subspace prediction components without retraining.
(iii) We evaluate CSP on frozen PIGNN-GC, GridSFM, gridfm-graphkit, and
LUMINA models on the 2224-bus GBnetwork, where it lowers AC power-balance
residuals and voltage-magnitude error in all four models and provides a
better initialization for one Newton update.
We further assess its
generality on 30 additional systems and its robustness under topology
and load-distribution shifts.

\section{Error Geometry and CSP}
\label{sec:method}

\subsection{AC power-balance consistency and calibration}

For an $N$-bus grid, let $x=[V,\theta]\in\mathbb{R}^{2N}$ denote the
voltage state, with $\underline V=V\odot e^{\jmath\theta}$.
Given operating condition $u$, a surrogate predicts $\hat{x}=f_\phi(u)$, from which $(P,Q)$
are recomputed using the network admittance model. For setpoints
$(P^{\mathrm{set}},Q^{\mathrm{set}})$, we define
\[
r(\hat x;u)=
\begin{bmatrix}
\Delta P\\ \Delta Q
\end{bmatrix}
=
\begin{bmatrix}
P^{\mathrm{set}}-P\\
Q^{\mathrm{set}}-Q
\end{bmatrix},
\qquad
\mathrm{PB}=\sqrt{\Delta P^2+\Delta Q^2}.
\]
PB uses $\Delta P$ at PV/PQ and $\Delta Q$ at PQ buses, zero elsewhere;
prescribed PV/slack magnitudes and the slack angle are restored before
scoring. Mean/Max PB aggregate all bus--scenario pairs on a common
100-MVA per unit (p.u.) base.

Before geometric analysis, we remove train-only prediction bias:
\[
\hat b=\mathbb{E}_{s\in\mathrm{tr}}[\hat x_s-x_s^\star],
\qquad
\hat x^C=\hat x-\hat b,
\]
where $s$ indexes scenarios, $\star$ denotes the converged NR reference,
and $C$ denotes calibration. $\hat b$ is frozen at test time; angle bias is estimated as the
per-bus circular mean of wrapped prediction errors.

\subsection{Error geometry and training solution subspace projection}

From training Newton--Raphson solutions, let
$X_V,X_\theta\in\mathbb{R}^{S_{\mathrm{tr}}\times N}$ stack centered
magnitudes and circularly centered angles:
\begin{equation}
X_V=W_V\Sigma_VU_V^\top,\quad
X_\theta=W_\theta\Sigma_\theta U_\theta^\top,\quad
P_k=U_kU_k^\top,
\end{equation}
where $U_k=\operatorname{diag}(U_{V,k},U_{\theta,k})$ contains the
leading $k$ state-space directions per channel; $k=16$ denotes 16 per channel.


For calibrated error $e=\hat x^C-x^\star$,
\begin{equation}
e^{\parallel}=P_ke,\quad
e^{\perp}=(I-P_k)e,\qquad
\hat x-x^\star=\hat b+e^{\parallel}+e^{\perp}.
\label{eq:error_decomposition}
\end{equation}
Here, $\hat b$ is the train-estimated mean bias, while
$e^{\parallel}$ and $e^{\perp}$ partition the remaining error into
in- and off-subspace components.

Around a converged state with fixed topology and bus types,
\begin{equation}
r(x^\star+e;u)
=
r(x^\star;u)+J(x^\star;u)e+\mathcal{O}(\|e\|_2^2),
\label{eq:local_residual}
\end{equation}
where $r(x^\star;u)=0$ and $J$ is the AC residual Jacobian; hence
equal-norm errors can yield different residuals by orientation.


Motivated by this orientational sensitivity, CSP first removes $\hat{b}$ and then projects onto the dominant training-solution directions:
\begin{equation}
\boxed{
\hat x^{\mathrm{CSP}}
=
\bar x_{\mathrm{tr}}
+
P_k\left(\hat x-\hat b-\bar x_{\mathrm{tr}}\right)
}
\label{eq:csp}
\end{equation}
where $\bar x_{\mathrm{tr}}$ is the training-solution mean;
both $\bar x_{\mathrm{tr}}$ and $P_k$ remain fixed at test time.
Projection can remove prediction error but also discard valid reference
variation. Let
$e^C=\hat x^C-x^\star$ and
$e^{\mathrm{CSP}}=\hat x^{\mathrm{CSP}}-x^\star$.
Since $P_k$ is orthogonal,
\begin{equation}
\|e^{\mathrm{CSP}}\|_2^2-\|e^C\|_2^2
=
\|(I-P_k)(x^\star-\bar x_{\mathrm{tr}})\|_2^2
-\|(I-P_k)e^C\|_2^2 .
\label{eq:tradeoff}
\end{equation}
Hence, CSP improves state accuracy when the removed off-subspace
prediction-error energy exceeds the discarded reference variation.

CSP is a lightweight, data-driven post-hoc output correction rather than a projection onto the nonlinear AC-feasible set; it therefore does not guarantee exact feasibility, but can improve physical consistency when the removed off-subspace components are residual-sensitive under the local AC Jacobian.



\section{Experiments and Results}
\label{sec:experiments}

Our primary study uses 37,022 converged pandapower Newton--Raphson
operating points on the 2224-bus GBnetwork, split 1:1:1 for
training/validation/test. The dataset combines global load/generation
scaling with independent load $P/Q$ and active-generation perturbations;
the backbone topology and PV-voltage control schedule remain fixed.
We evaluate PIGNN-GC, our global-context extension of
PIGNN-Attn-LS~\cite{kim2026pignn} retaining its line-search, GridSFM~\cite{yang2026gridsfm},
gridfm-graphkit~\cite{gridfmgraphkit2026}, and
LUMINA~\cite{li2026lumina}; GridSFM is also evaluated
cross-grid on 31 systems.

All models are trained from scratch using model-specific stable
initialization, and CSP is evaluated within each frozen model.
gridfm-graphkit uses its released architecture/configuration without a
pretrained checkpoint. Both GridSFM and LUMINA retain their released backbones and are adapted
from AC-OPF to AC-PF by fixing generator dispatch, predicting only voltage
states with PF-specific residuals, and applying known-variable masking.
All models use pre-solution pandapower~\cite{zimmerman2011matpower,thurner2018pandapower} internal PPC-derived inputs. Training and inference use one NVIDIA H100 96-GB GPU; inference timing uses the
largest memory-fitting batch to maximize GPU utilization. Calibration statistics and solution
bases use training data only. From $k\in\{4,8,16,32,64\}$, validation
Mean PB selects $k^\star=16$ without degrading calibrated
voltage-magnitude RMSE and is frozen for test and distribution-shift experiments.
Unless otherwise stated, mechanistic, ablation, and cross-grid analyses
use model seed 42, whereas Tables~1 and~3 aggregate seeds 41--43.

Voltage RMSE alone can be small even when a model largely predicts a
static per-bus profile rather than following scenario-dependent voltage
variation~\cite{yang2026gridsfm}. To isolate this scenario-dependent
response, we center each bus as
$\tilde{V}_{is}=V_{is}-\bar{V}_i$ and
$\tilde{\hat{V}}_{is}=\hat{V}_{is}-\bar{\hat{V}}_i$, and report
\begin{equation}
R_{\mathrm{w}}^2
=
\operatorname{corr}^2
\left(\tilde{V}_{is},\tilde{\hat{V}}_{is}\right),
\qquad
a_{\mathrm{w}}
=
\frac{\operatorname{Cov}
\left(\tilde{V}_{is},\tilde{\hat{V}}_{is}\right)}
{\operatorname{Var}\left(\tilde{V}_{is}\right)} .
\end{equation}
$R_{\mathrm{w}}^2$ is the squared correlation between the within-bus-centered
prediction and reference across all bus--scenario pairs, while
$a_{\mathrm{w}}$ measures the scale of the predicted variation
relative to the true variation.

\begin{table*}[t]
\centering
\caption{
GBnetwork results at $k=16$.
All models report mean $\pm$ standard deviation over three model seeds.
$C$ denotes calibration, $P_{16}$ projection, and CSP$_{16}$ their
composition. Inference time is GPU-synchronized forward execution plus
the corresponding output transform, reported per scenario.
}
\label{tab:intervention}
\setlength{\tabcolsep}{2.5pt}
\begin{tabular}{llrrrrrrr}
\toprule
Model & Output &
$V$ RMSE &
$\theta$ RMSE &
$R_{\mathrm{w}}^2 \uparrow$ &
$a_{\mathrm{w}}$ &
Mean PB &
Max PB &
Inference \\
& & $\downarrow$ & ($^\circ$) $\downarrow$ & & & $\downarrow$ & $\downarrow$ & (ms/scen.) \\
\midrule

Per-bus mean
& Baseline
& .02341
& 6.847
& .000
& .000
& .12084
& 5.88
& -- \\

\midrule
PIGNN-GC
& Raw
& $.02850{\pm}.00232$
& $1.108{\pm}.175$
& $.671{\pm}.050$
& $.546{\pm}.072$
& $.31592{\pm}.03240$
& $195.89{\pm}52.54$
& $10.034{\pm}.023$ \\

& $C$
& $.01393{\pm}.00105$
& $\mathbf{.777{\pm}.161}$
& $.671{\pm}.050$
& $.546{\pm}.072$
& $.16661{\pm}.02352$
& $190.00{\pm}37.53$
& $10.028{\pm}.023$ \\

& $P_{16}$
& $.01624{\pm}.00254$
& $1.091{\pm}.166$
& $\mathbf{.829{\pm}.045}$
& $.541{\pm}.071$
& $.07594{\pm}.00422$
& $5.07{\pm}.85$
& $10.036{\pm}.023$ \\

& CSP$_{16}$
& $\mathbf{.01222{\pm}.00134}$
& $.835{\pm}.141$
& $\mathbf{.829{\pm}.045}$
& $.541{\pm}.071$
& $\mathbf{.05497{\pm}.00257}$
& $\mathbf{3.93{\pm}.07}$
& $10.034{\pm}.032$ \\

\midrule
GridSFM
& Raw
& $.02868{\pm}.01652$
& $2.294{\pm}.302$
& $.768{\pm}.151$
& $.724{\pm}.149$
& $.26867{\pm}.13004$
& $9.10{\pm}4.02$
& $25.045{\pm}.207$ \\

& $C$
& $.01086{\pm}.00403$
& $\mathbf{1.781{\pm}.055}$
& $.768{\pm}.151$
& $.724{\pm}.149$
& $.08251{\pm}.01093$
& $7.29{\pm}3.53$
& $25.015{\pm}.096$ \\

& $P_{16}$
& $.01814{\pm}.01145$
& $2.244{\pm}.325$
& $\mathbf{.936{\pm}.032}$
& $.724{\pm}.149$
& $.07630{\pm}.01942$
& $5.08{\pm}.49$
& $25.087{\pm}.279$ \\

& CSP$_{16}$
& $\mathbf{.00782{\pm}.00307}$
& $1.807{\pm}.053$
& $\mathbf{.936{\pm}.032}$
& $.724{\pm}.149$
& $\mathbf{.05136{\pm}.00143}$
& $\mathbf{4.94{\pm}.42}$
& $24.577{\pm}.091$ \\

\midrule
gridfm-
& Raw
& $.00892{\pm}.00154$
& $.385{\pm}.042$
& $.905{\pm}.030$
& $.811{\pm}.045$
& $.09282{\pm}.01763$
& $19.82{\pm}8.65$
& $5.917{\pm}.019$ \\
graphkit

& $C$
& $.00753{\pm}.00118$
& $\mathbf{.367{\pm}.042}$
& $.905{\pm}.030$
& $.811{\pm}.045$
& $.07685{\pm}.01307$
& $18.88{\pm}8.69$
& $5.730{\pm}.004$ \\

& $P_{16}$
& $.00617{\pm}.00099$
& $.503{\pm}.018$
& $\mathbf{.959{\pm}.008}$
& $.801{\pm}.044$
& $.04593{\pm}.00156$
& $5.36{\pm}.56$
& $5.723{\pm}.011$ \\

& CSP$_{16}$
& $\mathbf{.00606{\pm}.00090}$
& $.496{\pm}.021$
& $\mathbf{.959{\pm}.008}$
& $.801{\pm}.044$
& $\mathbf{.04571{\pm}.00135}$
& $\mathbf{5.08{\pm}.18}$
& $5.779{\pm}.012$ \\

\midrule
LUMINA
& Raw
& $.02653{\pm}.00031$
& $2.703{\pm}.313$
& $.443{\pm}.014$
& $.346{\pm}.012$
& $1.29896{\pm}.85056$
& $664.19{\pm}448.00$
& $7.548{\pm}0.369$ \\

& $C$
& $.01780{\pm}.00020$
& $1.442{\pm}.060$
& $.443{\pm}.014$
& $.346{\pm}.012$
& $.21432{\pm}.01186$
& $20.65{\pm}15.82$
& $7.380{\pm}0.155$ \\

& $P_{16}$
& $.01898{\pm}.00049$
& $2.160{\pm}.268$
& $\mathbf{.906{\pm}.012}$
& $.338{\pm}.012$
& $.07641{\pm}.00128$
& $5.47{\pm}.56$
& $7.981{\pm}0.868$ \\

& CSP$_{16}$
& $\mathbf{.01570{\pm}.00024}$
& $\mathbf{1.282{\pm}.058}$
& $\mathbf{.906{\pm}.012}$
& $.338{\pm}.012$
& $\mathbf{.06662{\pm}.00074}$
& $\mathbf{4.61{\pm}.19}$
& $7.891{\pm}0.567$ \\

\bottomrule
\end{tabular}
\end{table*}

\subsection{Off-subspace error and directional sensitivity.}

\noindent\textit{Off-subspace error.}
Off-subspace error is substantial across models and is positively associated
with Mean PB. Although the rank-16 basis explains 98.8\%/99.7\% of training
voltage-magnitude/angle variation, the median scenario-wise fractions of total squared
error lying outside the respective subspaces are
26.3/55.0/56.8/27.1\% for magnitude and
10.3/2.7/39.6/31.7\% for angle, for
PIGNN-GC/GridSFM/gridfm-graphkit/LUMINA, respectively.
After accounting for differences in total error norms, scenarios with larger
off-subspace fractions have higher Mean PB, with partial Spearman correlations
of .659/.635/.594/.833 for magnitude and
.789/.906/.809/.860 for angle (all 95\% CIs exclude zero).
\vspace{0.35em}

\begin{figure}[t]
    \centering
    \includegraphics[width=\columnwidth]
    {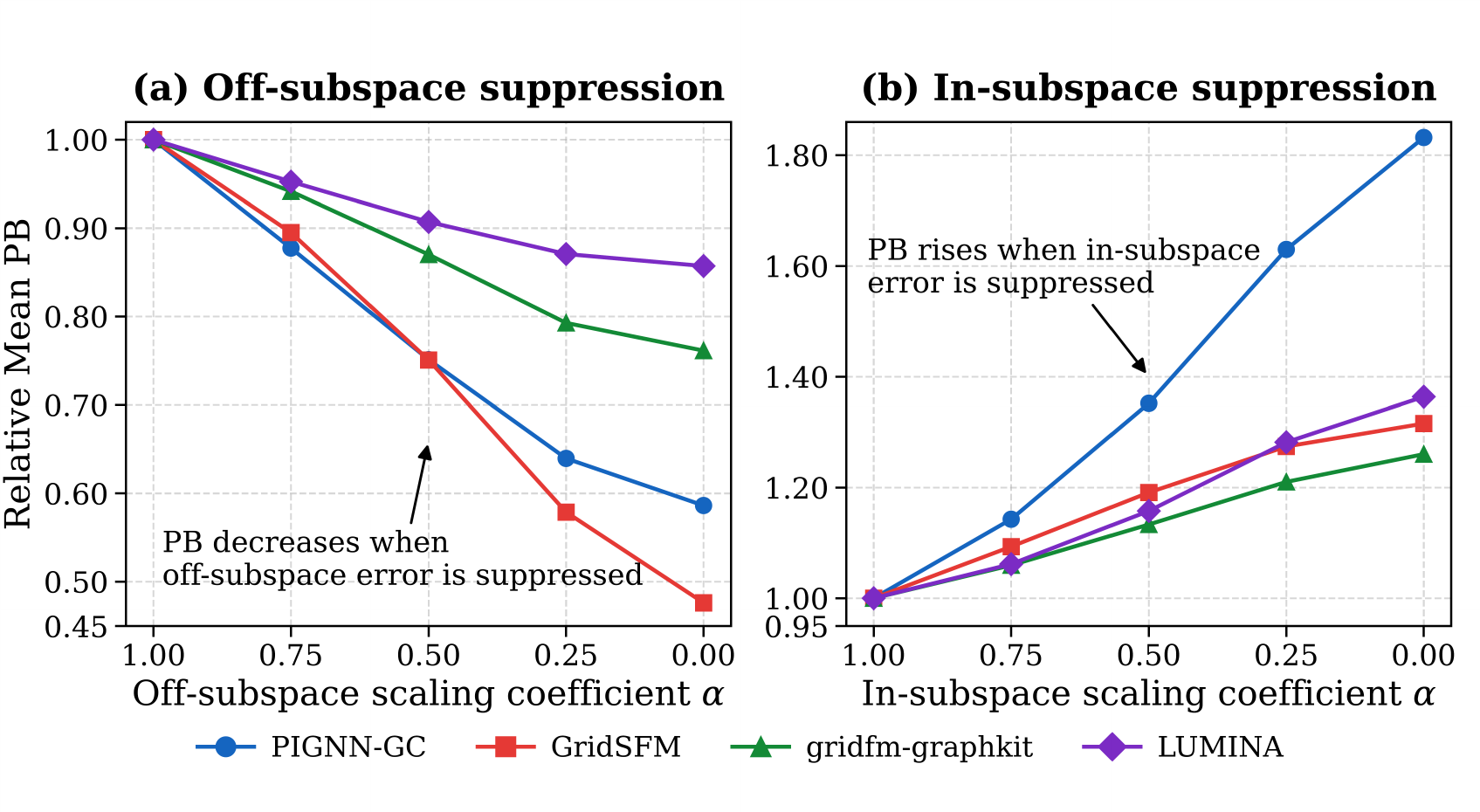}
    \caption{
    Fixed-norm directional intervention on GBnetwork, with Relative Mean PB
    normalized at $\alpha=1$.
    }
    \label{fig:error_geometry}
\end{figure}

\noindent\textit{Fixed-norm intervention.}
At fixed error norm, suppressing off-subspace error consistently lowers
Relative Mean PB across all four surrogates, whereas suppressing in-subspace
error has the opposite effect (Fig.~\ref{fig:error_geometry}), isolating
error orientation from error magnitude. We test this with an oracle
diagnostic using the known reference state, varying the voltage-magnitude
error orientation at fixed $\ell_2$ norm:
\begin{equation}
\tilde e(\alpha)
=
\|e_V\|_2
\frac{e_V^\parallel+\alpha e_V^\perp}
{\|e_V^\parallel+\alpha e_V^\perp\|_2},
\label{eq:controlled_geometry}
\end{equation}
where phase angles remain at their calibrated predictions. Decreasing $\alpha$ suppresses off-subspace error; the control analogously
suppresses the in-subspace component by exchanging
$e_V^\parallel$ and $e_V^\perp$.
\vspace{0.35em}

\noindent\textit{Jacobian sensitivity.}
Off-subspace directions are substantially more residual-sensitive than
in-subspace directions under the local AC Jacobian, consistent with the
fixed-norm intervention. Using directional gain
$g(d)=\|Jd\|_2/\|d\|_2$, 32 random direction pairs at each of 128 held-out
operating points yield a median paired off-/in-subspace gain ratio of
$1{,}563\times$. The same gain contrast appears in the actual calibrated
prediction errors: the ratios of median off- to in-subspace gains are
$22.81\times$, $575.83\times$, $4.69\times$, and $6.51\times$ for
PIGNN-GC, GridSFM, gridfm-graphkit, and LUMINA, respectively.
The observed error regime is consistent with the first-order expansion in
Eq.~\eqref{eq:local_residual}, with median linearized-to-nonlinear
residual-change ratios between $0.996$ and $1.009$ across all four models.

\subsection{Solution subspace specificity.}
Our matched-basis comparison shows that preserving the same amount of
reference-state variation is insufficient for low PB, indicating that the
specific state-space directions retained by the basis are important.
We compare SVD$_{16}$ with graph-Laplacian and spatial low-pass controls,
each tuned using training references to match the $V/\theta$ variation
retained by SVD$_{16}$. On the test set, all three preserve nearly identical
reference variation (Table~\ref{tab:basis_control}), yet the alternatives yield
substantially larger PB for both projected references and corrected predictions.




\begin{table}[t]
\centering
\caption{
Matched basis comparison on GBnetwork for PIGNN-GC.
Parentheses denote $V/\theta$ ranks or smoothing strengths.
}
\label{tab:basis_control}
\setlength{\tabcolsep}{1.8pt}
\begin{tabular}{lrrr}
\toprule
Basis &
Ref. var. &
Ref. &
Pred. \\
& retained (\%) & PB & PB \\
\midrule
SVD ($16/16$)
& $98.789/99.683$
& \textbf{.0415}
& \textbf{.0569} \\

Laplacian matched ($1750/2122$)
& $98.819/99.688$
& 26.15
& 26.22 \\

Spatial matched ($\beta=.357/.208$)
& $98.801/99.684$
& 6.20
& 6.22 \\
\bottomrule
\end{tabular}
\end{table}

\subsection{Post-training CSP correction}

Table~\ref{tab:intervention} compares Raw, $C$, $P_{16}$, and CSP$_{16}$ across the four frozen surrogates. Relative to $C$, CSP reduces Mean
PB by $67.0\%$, $37.8\%$, $40.5\%$, and $68.9\%$, respectively, while
lowering voltage-magnitude RMSE in all four models.
Projection improves $R_w^2$ with little change in $a_w$; angle
accuracy is model dependent. Raw and CSP16 inference times are
comparable.

Magnitude-only, angle-only, and full projection show model-dependent effects of
angle projection. PIGNN-GC achieves the lowest PB with full CSP, whereas
GridSFM performs best with magnitude-only projection. For gridfm-graphkit,
adding angle projection reduces Mean/Max PB from $.0612/18.23$ to $.0457/5.08$,
but degrades angle RMSE from $.367^\circ$ to $.496^\circ$. For LUMINA, it
reduces Mean/Max PB from $.1939/38.53$ to $.0660/4.51$ while improving angle RMSE
from $1.374^\circ$ to $1.215^\circ$. Thus, better physical consistency from
angle projection need not imply better angle accuracy.

These model-dependent accuracy changes are explained by
Eq.~\eqref{eq:tradeoff}. For magnitude, the removed off-subspace
prediction-error energy exceeds the discarded reference variation
across all four models. For angle, the balance differs across models:
in gridfm-graphkit, angle projection discards more valid reference
variation than model error and therefore worsens angle RMSE, whereas
the opposite holds for LUMINA.

\begin{figure}[t]
    \centering
    \includegraphics[width=1.0\columnwidth]{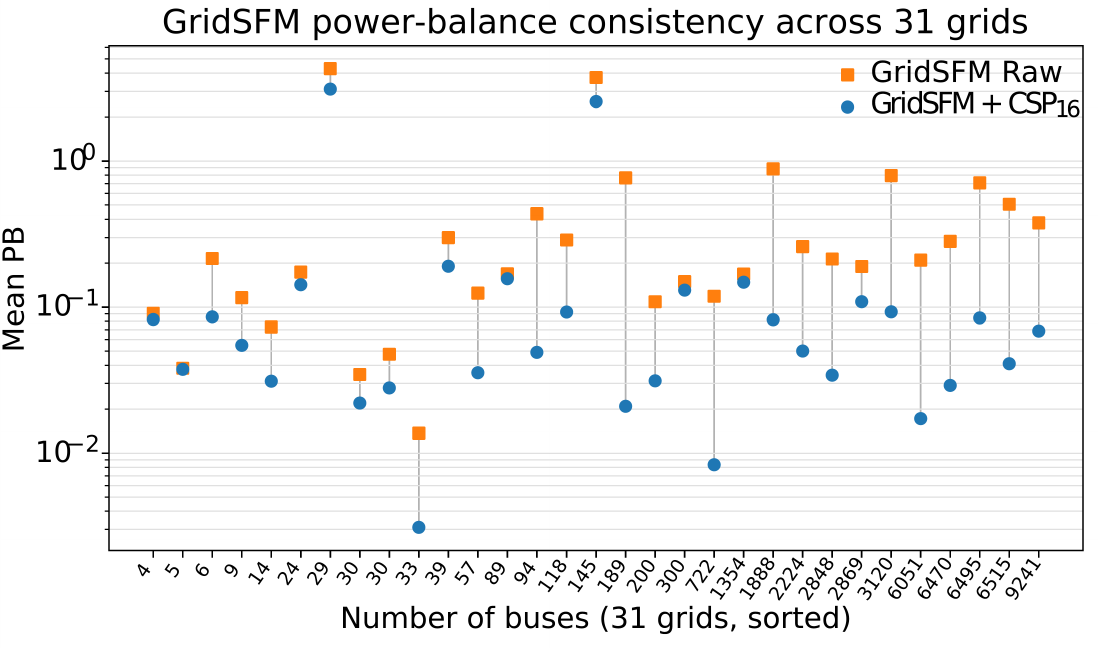}
    \caption{
    Grid-by-grid physical consistency of GridSFM before and after
    CSP$_{16}$ across 31 systems ordered by bus count. Mean PB is shown
    on a logarithmic scale.
    }
    \label{fig:crossgrid_csp}
\end{figure}

\subsection{Complementarity with one-step Newton refinement.}

CSP provides a consistently better initialization for one-step Newton
refinement, improving both state accuracy and post-update physical
consistency across all four models. Starting from $C$ or CSP$_{16}$, we
apply one Newton update ($\mathrm{NR}_1$) to non-slack angles and PQ voltage
magnitudes, with validation-selected damping factor; $\eta=1$ is selected for all models, with no linear-solve failures. Relative to $C$
initialization, CSP lowers magnitude and angle RMSE in all models and
reduces post-update Mean PB by $21.1$--$96.4\%$
(Table~\ref{tab:nr1}).

The benefit is particularly pronounced for LUMINA: $C+\mathrm{NR}_1$
shows a large one-seed overshoot despite a successful linear solve, whereas
CSP$_{16}$ initialization remains consistent across seeds.

\begin{table}[t]
\centering
\caption{
One Newton update from calibrated or CSP$_{16}$ outputs on GBnetwork.
$V$ RMSE and Mean PB are reported in $10^{-3}$.
Values are means $\pm$ standard deviations over three model seeds.
}
\label{tab:nr1}
\setlength{\tabcolsep}{2.2pt}
\begin{tabular}{llrrr}
\toprule
Model & Start &
$V$ RMSE &
$\theta$ RMSE &
Mean PB \\
& & $(10^{-3})$ & ($^\circ$) & $(10^{-3})$ \\
\midrule

PIGNN-GC
& $C$
& $1.80{\pm}.12$
& $.119{\pm}.013$
& $2.84{\pm}.46$ \\
& CSP$_{16}$
& $\mathbf{1.17{\pm}.10}$
& $\mathbf{.073{\pm}.009}$
& $\mathbf{1.48{\pm}.13}$ \\

\midrule
GridSFM
& $C$
& $11.86{\pm}6.77$
& $.681{\pm}.380$
& $2.50{\pm}1.54$ \\
& CSP$_{16}$
& $\mathbf{.97{\pm}.20}$
& $\mathbf{.059{\pm}.013}$
& $\mathbf{1.22{\pm}.29}$ \\

\midrule
gridfm-
& $C$
& $.943{\pm}.232$
& $.0843{\pm}.0236$
& $.941{\pm}.325$ \\
graphkit & CSP$_{16}$
& $\mathbf{.496{\pm}.085}$
& $\mathbf{.0277{\pm}.0041}$
& $\mathbf{.742{\pm}.080}$ \\

\midrule
LUMINA
& $C$
& $13.71{\pm}9.23$
& $.671{\pm}.038$
& $46.08{\pm}64.15$ \\
& CSP$_{16}$
& $\mathbf{1.040{\pm}.016}$
& $\mathbf{.0802{\pm}.0031}$
& $\mathbf{1.653{\pm}.036}$ \\

\bottomrule
\end{tabular}
\end{table}
\subsection{Cross-grid breadth and distribution shift}
\label{sec:breadth_shift}

\noindent\textit{Cross-grid breadth.}
CSP reduces GridSFM Mean PB across all 31 systems
(Fig.~\ref{fig:crossgrid_csp}), spanning 4--9,241 buses and sub-kV to
750-kV levels. The set includes 28 pandapower/PYPOWER-format systems
from IEEE, PEGASE, RTE, and GB benchmark families, plus three Common
Information Model (CIM)-based networks. For each system, calibration and
subspace statistics are fitted from its training data, using nominal
$k=16$ per channel and the available SVD components on smaller systems.
PV-voltage setpoints vary for most systems, but are fixed for GBnetwork with conflicting controllers.

\vspace{0.35em}

\noindent\textit{Topology and spatial shifts.}
CSP remains stable in most statistics under $N$--1/$N$--2 topology
shifts, with degradation concentrated in isolated worst-case violations,
whereas GBcorr causes a broader performance drop.
Under the topology shifts, voltage accuracy, tracking, and bulk PB remain
close to control, while larger maxima motivate worst-case assessment beyond
average error. GBcorr applies correlated, spatially
varying load scaling across 14 zones while preserving total demand and
generation settings. Transferring both calibration and basis gives Mean PB of
$0.1003$ p.u.; refitting calibration alone reduces this to $0.0944$,
basis alone to $0.0740$, and both to $0.068$.
Although the transferred/refitted bases retain $93.55\%/98.53\%$ of GBcorr
magnitude variation, projecting converged GBcorr references yields Mean PBs
of $0.0868/0.0438$ p.u., respectively. This reference-only diagnostic
isolates the effect of basis transfer from surrogate error and helps explain
the improvement from basis refitting.

\begin{table}[t]
\centering
\caption{
PIGNN-GC with fixed-rank CSP$_{16}$ under topology and
operating-distribution shifts. T transfers the control calibration and basis,
whereas R refits both on condition-matched training data.
}
\label{tab:shift}
\setlength{\tabcolsep}{1.2pt}
\begin{tabular}{lrrrrrrrr}
\toprule
Condition &
\(V\) RMSE &
$R_{\mathrm{w}}^2 \uparrow$ &
$a_{\mathrm{w}}$ &
Mean &
Med. &
p95 &
p99 &
Max \\
\midrule
Control
& .01281
& .8476
& .4866
& .0581
& .0131
& .2665
& .5462
& 3.58 \\
\midrule
$N$--1 (T)
& .01314
& .8391
& .4750
& .0603
& .0131
& .2713
& .5632
& 33.16 \\
$N$--2 (T)
& .01288
& .8458
& .4870
& .0617
& .0132
& .2713
& .5721
& 27.45 \\
\midrule
GBcorr (T)
& .01966
& .4819
& .4025
& .1003
& .0231
& .4512
& .9746
& 7.00 \\
GBcorr (R)
& \textbf{.01620}
& .5150
& .4247
& \textbf{.0680}
& .0174
& .3031
& \textbf{.6415}
& \textbf{4.99} \\
\bottomrule
\end{tabular}

\end{table}

\section{CONCLUSION}

This work identifies output-error geometry as an important factor in the gap
between voltage accuracy and physical consistency in neural AC
power-flow surrogates. Across four architecturally diverse neural surrogates, we show that small voltage errors can contain off-subspace
components that induce large AC power-balance violations. Building on
this insight, calibrated solution-subspace projection suppresses these
components, reducing voltage-magnitude error and power-balance residuals
without retraining; the residual reduction extends to 30 additional
grids. These results suggest that error direction, not only magnitude,
should inform surrogate evaluation and design. They also motivate
extending physics-consistent neural surrogates toward broader
power-system analysis, estimation, and optimization.

\bibliographystyle{IEEEbib}
\bibliography{neural_pf_tracking_calibration_error_geometry}

\end{document}